\documentclass[10pt,twocolumn]{article}

\usepackage[margin=0.75in]{geometry}
\usepackage{amsmath, amssymb}
\usepackage{graphicx}
\usepackage{booktabs}
\usepackage{cite}
\usepackage{hyperref}
\usepackage{caption}
\usepackage{cite}
\usepackage{algorithm}
\usepackage{algpseudocode}
\usepackage{multirow}

\title{A Kinematic Framework for Screening Candidate Pinch Configurations in Robotic Hand Design without Object or Contact Models}

\author{
    HyoJae Kang$^{1}$, Joonho Lee$^{1}$, Hyunmok Jung$^{1}$, and Dong Il Park$^{1,*}$ \\
    $^{1}$Advanced Robotics Research Center, Korea Institute of Machinery \& Materials (KIMM) \\
    $^{*}$Corresponding author
}

\begin{document}
\date{}
\maketitle

% ======================
% Footnote (중요)
% ======================
\thispagestyle{plain}
\footnotetext{This manuscript has been accepted for publication in IEEE Access. DOI: 10.1109/ACCESS.2026.3729781}

% ======================
\begin{abstract}
Evaluating the pinch capability of a robotic hand is important for understanding its functional dexterity. However, many existing grasp evaluation methods rely on object geometry or contact force models, which limits their applicability during the early stages of robotic hand design. This study proposes a kinematic evaluation method for analyzing pinch configurations of robotic hands based on interactions between fingertip workspaces. First, the reachable workspace of each fingertip is computed from the joint configurations of the fingers. Then, feasible pinch configurations are detected by evaluating the relationships between fingertip pairs. Since the proposed method does not require information about object geometry or contact force models, the pinch capability of a robotic hand can be evaluated solely based on its kinematic structure. In addition, analyses are performed on four different kinematic structures of the hand to investigate their impact on the pinch configurations. The proposed evaluation framework can serve as a useful tool for comparing different robotic hand designs and analyzing pinch capability during the design stage.
\end{abstract}

% ======================
\section{Introduction}
The human hand is one of the body’s most functional and essential parts, capable of carrying out a wide range of daily tasks \cite{c1,c2}. It is considered one of the most complex and sophisticated tools that humans rely on for daily tasks \cite{c3}. Inspired by the versatility of the human hand, robotic systems have been equipped with various end-effectors designed to perform similar functions \cite{c4}. Robotic hands have been widely utilized to replace human workers in heavy, repetitive, and hazardous working environments \cite{c5,c6,c7}.

To perform such tasks, a wide range of robotic grippers and hands have been developed, ranging from simple two-finger grippers with low degrees of freedom (DoF) to multi-finger robotic hands and anthropomorphic hands that resemble the human hand structure. Recently, the demand for robotic systems capable of operating in human environments has increased \cite{c202}. These systems are expected to perform collaborative tasks with humans and grasp objects with various shapes and sizes \cite{c8}. As a result, the development of robotic hands with enhanced grasping versatility and adaptability has become an important research topic.

To handle a wide variety of objects, robotic hands require a high level of dexterity. Dexterous robotic hands aim to achieve grasping and manipulation capabilities similar to those of the human hand. Dexterity has been widely used to describe the skillfulness of robotic equipment, particularly in robotic hands. Previous studies have categorized dexterity evaluation methods into several groups based on the aspects of performance they measure \cite{c9}. One category is potential dexterity, which evaluates the set of reachable hand configurations without considering object interaction. Potential dexterity reflects the diversity of possible hand postures and the inherent capability of the hand mechanism. This category includes measures such as kinematic redundancy and thumb opposability. Kinematic redundancy refers to the availability of multiple joint configurations that can achieve the same end-effector state, while thumb opposability evaluates the ability of the thumb to interact with other fingers. Thumb opposability is commonly assessed using methods such as workspace overlap between the thumb and other fingers \cite{c10,c11} or the Kapandji test \cite{c12}.

Another category is grasp dexterity, which evaluates the set of stable hand configurations when grasping an object. This approach is often based on grasp taxonomy, which classifies different grasp types according to the contact relationships between the hand and the object \cite{c9}. Grasp taxonomy provides insight into the variety of grasp strategies that a robotic hand can achieve when interacting with objects. The third category is manipulability dexterity, which evaluates the ability to change the position and orientation of a grasped object within the workspace while maintaining stable contact \cite{c1}. This concept is commonly used to assess the capability of robotic hands to perform object manipulation tasks after a stable grasp has been established.

In the design of robotic hands, particularly anthropomorphic hands, dexterity is often considered an important performance indicator because it is closely related to the potential capability of the hand to perform various tasks. Evaluating dexterity during the design stage can therefore help predict the performance of a robotic hand before it is physically constructed. While some robotic hands are designed for specific tasks, many designs aim to achieve general-purpose functionality. In such cases, it is difficult to evaluate the expected task performance in advance because the objects to be manipulated and the corresponding task trajectories are not explicitly defined. Furthermore, many evaluation methods require assumptions about contact models between the hand and the object, which makes it challenging to apply them during the early stages of hand design when such information is unavailable.

While these dexterity evaluation methods provide useful insights into the general capabilities of robotic hands, they do not directly indicate whether specific grasp configurations can be formed. In practical manipulation tasks, the ability to realize particular grasp types is often more relevant than the overall variety of hand postures. Among various grasp types, pinch grasps are widely used for precise object handling and frequently appear as a fundamental grasp configuration. Therefore, analyzing the capability of a robotic hand to form pinch configurations can provide useful insight into its potential grasping functionality.

Motivated by this observation, this study focuses on evaluating pinch configurations in a five-finger robotic hand. In particular, this study proposes a method to analyze four types of pinch grasp based on the reachable hand postures obtained through kinematic analysis. Since the evaluation is performed using reachable configurations of the hand, the proposed method does not rely on force or torque models. Therefore, the proposed approach allows the pinch capability of a robotic hand to be evaluated solely from its kinematic structure. This makes the method particularly suitable for the early design stage, where detailed information about objects and contact conditions is typically unavailable. In practical grasping situations, many factors such as grasping force, object shape, size, weight, and friction coefficient affect the grasp stability. However, these factors are excluded in order to focus on a purely kinematic analysis of hand postures. This approach allows the structural characteristics of the robotic hand to be evaluated during the design stage without requiring detailed information about the object or task. The main contributions of this study are summarized as follows:

\begin{enumerate}
    \item A kinematic evaluation method for analyzing pinch configurations with respect to four types of pinch grasp.
    \item Quantitative analysis of pinch configurations for anthropomorphic five-finger hands with different degree-of-freedom configurations.
    \item Comparative evaluation of thumb–finger pinch enabled by thumb opposability and non-thumb pinch configurations between other fingers.
\end{enumerate}

The remainder of this paper is organized as follows. Section 2 introduces the five-fingered hands and the types of pinch grasp considered in this study, as well as the evaluation methods. Section 3 describes the kinematic structures of the four hands analyzed in this work. Section 4 presents three methods for evaluating pinch grasp. Section 5 analyzes the reachable configurations of the hands and presents the evaluation results. Finally, Section 6 summarizes the conclusions and contributions of this study, and Section 7 discusses the limitations of the work and directions for future research.

%%%%%%%%%%%%%%%%%%%%%%%%%%%%%%%%%%%%%%%%%%%%%%%%%%%%%%%%%%%%%%%%%%%%%%
\section{Related Works}

Multi-finger robotic hands have been extensively studied to achieve versatile grasping and manipulation capabilities. Compared with simple two-finger grippers, multi-finger hands provide richer contact combinations and enable a wider range of grasp configurations. Various five-finger robotic hands have been developed in previous works \cite{c14,c97,c16,c91,c18,c96,c98,c20,c93,c90,c99,c94,c95,c100}, differing in the number of DoF assigned to each finger and thumb in Table ~\ref{tab:1}. Typically, finger joints provide flexion/extension(F/E) and abduction/adduction(A/A) motions, while the thumb includes additional opposition/reposition(O/R) motion that enables interaction with other fingers. These structural differences influence the reachable contact regions between fingers and affect the types of grasp configurations that can be achieved. 

\begin{table}[h]
\caption{Comparison of finger and thumb DoF in existing robotic hands}
\centering{%
\begin{tabular}{c c c}
\toprule
\multirow{2}{*}{Description} & \multicolumn{2}{c}{DoF} \\
 &  (Finger) & (Thumb)\\
\midrule
Mini X \cite{c14}, IH2 Azzurra \cite{c97} & 2 & 3 \\
\midrule
ILDA \cite{c16}, HRI hand \cite{c91} & 3 & 3 \\
\midrule
Manopla \cite{c18}, Handroid \cite{c96} &  \multirow{2}{*}{3} & \multirow{2}{*}{4}\\
ORCA \cite{c98} &  &  \\
\midrule
MCR Hand III \cite{c20}  & \multirow{4}{*}{4} & \multirow{4}{*}{4}\\
Gifu Hand III \cite{c90} &  &  \\
HX5-D20 \cite{c93} & & \\
RAPID Hand \cite{c99} & & \\
\midrule
Wave \cite{c94}, Shadow \cite{c95} & \multirow{2}{*}{4} & \multirow{2}{*}{5}\\
MM Hand 1.0 \cite{c100} &  &  \\
\bottomrule
\end{tabular}
}%
\label{tab:1}
\end{table}

Pinch grasp plays an important role in precise object handling. In pinch grasp, objects are typically held between the thumb and one or more opposing fingers, enabling accurate positioning and stable contact control. The ability of the thumb to oppose other fingers is considered one of the key characteristics of the human hand. These approaches provide insight into the potential interaction between the thumb and other fingers but do not directly evaluate specific pinch configurations that may occur during grasping tasks. 

From the perspective of grasp function, pinch techniques have also been classified into several representative categories. Previous research \cite{c21} classifies pinch techniques into four categories: pulp pinch, tip pinch, lateral pinch, and three-jaw chuck pinch. A lateral pinch refers to a grasp in which force is applied between the radial side of the index finger’s middle phalanx and the pad of the thumb\cite{c22}. A pulp pinch is defined as a pinch generated between the pad of another finger and  the pad of the thumb \cite{c23}. Depending on which finger is involved, pulp pinches can be further classified as pulp-2, pulp-3, up to pulp-5 pinch \cite{c24}. A tip pinch occurs when the tip of the thumb opposes the tip of another finger to generate the pinching force \cite{c25}. In a three-jaw chuck pinch, an object is positioned between the tip of the distal phalanx of the fingers and the pad of the thumb. This grasp is closely related to the pulp-to-pulp pinch, in which the pad of the thumb’s distal phalanx contacts the pad of the distal phalanx of the fingers\cite{c26}. These human pinch types provide useful references for designing robotic pinch configurations.

Although these classifications describe different functional forms of pinch grasp, they do not provide a systematic method for evaluating whether a robotic hand can realize these pinch interactions. Existing approaches commonly focus on kinematic dexterity, grasp taxonomy, or manipulability indices to assess the overall performance of robotic hands. However, these methods do not explicitly evaluate pinch configurations between specific fingers. 

If the reachable area is unknown, infeasible grasps may be attempted, or the evaluation may require a long computation time \cite{c201}. While identifying the reachable area is important, it is also crucial to distinguish the feasible regions for each grasp configuration. Therefore, a quantitative evaluation framework for robotic pinch configurations is required. In particular, systematic evaluation methods comparing pinch configurations involving thumb opposition and those without the thumb remain limited.

Therefore, this study proposes an evaluation framework for pinch configurations in a five-finger robotic hand. The proposed framework evaluates representative pinch interactions based on reachable configurations. First, distal phalanx alignment is analyzed to evaluate pulp pinch and three-jaw chuck pinch, as well as pinching interactions that occur among the fingers without the involvement of the thumb. Second, lateral pinch capability is evaluated by analyzing the posture and distance between the thumb pad and the radial side of the index finger. Specifically, the feasible area for the lateral pinch is extended laterally, not only at the middle phalanx but across all phalanges. Third, tip pinch capability is assessed based on the distance between fingertip contact points. The tip pinch is considered not only between the thumb and the index finger but also in relation to the thumb and the other fingers. Through these evaluations, the proposed method provides a quantitative analysis of pinch interaction capabilities in robotic hands.

%%%%%%%%%%%%%%%%%%%%%%%%%%%%%%%%%%%%%%%%%%%%%%%%%%%%%%%%%%%%%%%%%%%%%%
\section{Kinematic Structures}
In this section, the kinematic structures of the five-fingered hand are introduced. The hand configuration is crucial for assessing grasp success, robustness, intended object pose, and even the shape of the grasped object \cite{c200}. Therefore, in this study, evaluations are conducted for a total of four different hand configurations.

\begin{figure*}[!t]
    \centering
    \includegraphics[width=0.91\textwidth]{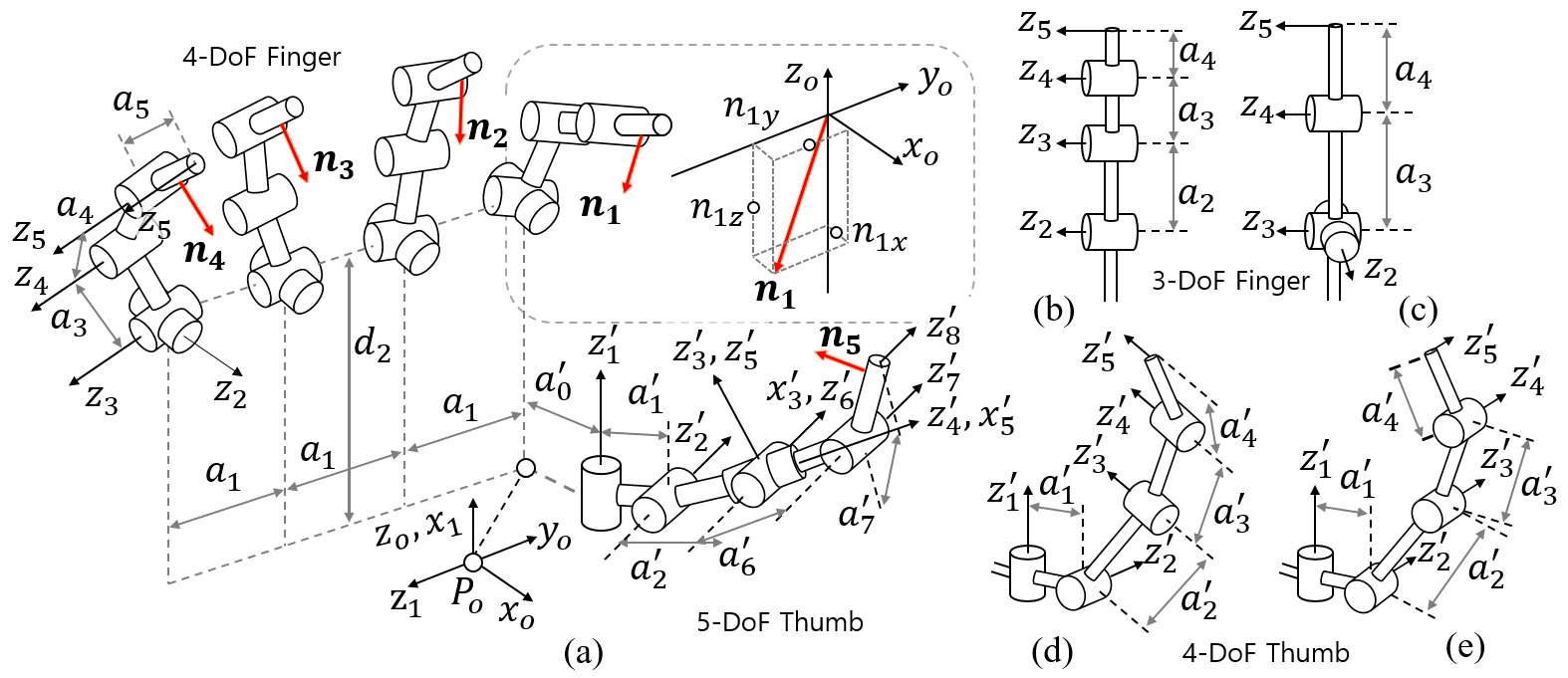}
    \caption{Kinematic structures of robotic hand (a) 21-DoF hand, (b) Cases of 3-DoF finger, (c) Cases of 4-DoF thumb}
    \label{fig:1}
\end{figure*}

First, based on Table ~\ref{tab:1}, the ranges of DoF for the finger and the thumb can be identified. Based on these ranges, the possible kinematic structures are illustrated in Fig. ~\ref{fig:1}. First, the hand illustrated in Fig. ~\ref{fig:1}(a) consists of four fingers and one thumb. Each finger has four DoF, enabling three DoF of F/E motion and one DoF of A/A motion. The thumb has five DoF, enabling three DoF of F/E motion, one DoF of A/A motion, and one DoF of O/R motion.

The DoF of the finger listed in Table ~\ref{tab:1} range from two to four. However, in the case composed of only three DoF of F/E motion, as shown in Fig. ~\ref{fig:1}(b), this configuration can be represented by removing the A/A actuation angle from the finger shown in Fig. ~\ref{fig:1}(a). In addition, when the A/A motion is removed from the three DoF finger composed of two DoF of F/E motion and one DoF of A/A motion, as shown in Fig. ~\ref{fig:1}(c), it becomes equivalent to a two DoF finger. The DoF of the thumb listed in Table ~\ref{tab:1} range from three to five. However, the case implemented by removing the offset and suppressing the actuation angle from the five DoF thumb shown in Fig. ~\ref{fig:1}(a), as illustrated in Fig. ~\ref{fig:1}(e), was excluded from the analysis. Fig. ~\ref{fig:1}(d) shows a case in which the motion is realized through a different arrangement of DoF. In this study, the considered finger structures are the four DoF finger (Fig. ~\ref{fig:1}(a)) and the three DoF finger (Fig. ~\ref{fig:1}(c)). For the thumb, the considered structures are the five DoF thumb (Fig. ~\ref{fig:1}(a)) and the four DoF thumb (Fig. ~\ref{fig:1}(d)). Therefore, a total of four combinations are considered in the analysis. 

Next, the kinematic parameters corresponding to the combinations considered in this study are introduced. Since the structure shown in Fig. ~\ref{fig:1}(a) has the largest number of DoF, the parameters of the other configurations were determined based on the values of this structure. It should also be noted that more variations can be considered depending on the kinematic parameters, offsets, and initial installation angles.

First, a coordinate frame $P_o x_o y_o z_o$ is defined with respect to the origin $P_o$ indicated in Fig. ~\ref{fig:1}(a). The offset from $P_o$ to the location corresponding to the first DoF of the thumb is denoted as $a_1'$. The lengths between the finger joints are denoted as $a_2'$, $a_3'$, $a_5'$, and $a_6'$. For structural simplification, the $x$-axes of $z_1'$ and $z_2'$ are parallel to the $x_o$ axis, and there is no offset along the $z_o$ direction. For the four DoF finger shown on the left side of Fig. ~\ref{fig:1}(c), the $x_1'$ axis and the $z_2'$ axis are parallel to the $x_o$ axis. In addition, the $z_i'$ axes ($i=1$ to end) represent the actuation axes, and the actuation angle at each axis is denoted by $\theta_i'$.

Next, for the fingers, the index finger is installed at a position located at a distance $d_1$ from the origin $P_o$ along the $z_o$ direction. The middle, ring, and little fingers are also located at the same distance. In addition, these fingers are arranged at equal intervals of $a_2$ along the $-y_o$ direction with respect to the index finger. All fingers have identical dimensions, and the lengths of the finger segments are $a_4$, $a_5$, and $a_6$. In the case shown on the right side of Fig. ~\ref{fig:1}(b), the fingers are installed at the same locations, and the lengths of the finger segments are $a_4$ and $a_5$. The $z_i$ axes ($i=2$ to end) correspond to the actuation axes, and the actuation angle at each axis is denoted by $\theta_i$. The motion configurations for the four cases considered are summarized in Table ~\ref{tab:2}.

\begin{table}[h]
    \caption{DoF configuration of the finger and thumb for each case}
    \centering
    \begin{tabular}{c|c c|c c}
    \toprule
    Cases & \multicolumn{2}{c}{Finger} & \multicolumn{2}{c}{Thumb}\\
    & 3-DoF & 4-DoF & 4-DoF & 5-DoF\\
    \midrule
    1 & O & X & O & X \\
    2 & O & X & X & O \\
    3 & X & O & O & X \\
    4 & X & O & X & O \\
    \bottomrule
    \end{tabular}
\label{tab:2}
\end{table}

First, the kinematic parameters of Case 4, which has the largest number of DoF, are considered. For the lengths of the finger segments, the index finger length ratios reported in \cite{c27} were referenced. Let the total finger length in Fig. ~\ref{fig:1}(a) be denoted as $l_f$. The length ratios of each segment are then given as follows:

\begin{equation}
a_4 = 0.51 l_f \quad\textrm{and}\quad a_5 = 0.27 l_f \quad\textrm{and}\quad a_6 = 0.22 l_f.
\end{equation}

The segment length ratios of the thumb were determined based on \cite{c28}. Let the total thumb length in Fig. ~\ref{fig:1}(a) be denoted as $l_t$. The length ratios of each segment are given as follows:

\begin{equation}
    a_2' = 0.47 l_t \quad\textrm{and}\quad a_6' = 0.31 l_t \quad\textrm{and}\quad a_7' = 0.22 l_t.
\end{equation}

Let the distance from the index finger to the little finger be defined as the hand width (HW). Based on the ratio between the hand width and overall hand length (HL) presented in \cite{c30}, the average value is determined as follows:

\begin{equation}
    HW = 3a_2 \quad\textrm{and}\quad HL = d_1 + l_f \quad\textrm{and}\quad HW = 0.54 HL.
\end{equation}

\begin{table}[h]
    \caption{Kinematic parameter ratios based on the hand length - identical for four cases}
    \centering
    \begin{tabular}{c|c|c|c|c|c|c}
    \toprule
    HL & HW & $l_f$ & $l_t$ & $a_1$ & $a_0'$ & $a_1'$\\
    \midrule
    1 & 0.54 & 0.45 & 0.51 & 0.18 & 0.1 & 0.1\\
    \bottomrule
    \end{tabular}
\label{tab:3}
\end{table}

\begin{table}[h]
    \caption{Kinematic parameter ratios based on the hand length for four cases}
    \centering
    \begin{tabular}{c|c|c|c|c|c|c}
    \toprule
    Case & $a_3$ & $a_4$ & $a_5$ & $a_2'$ & $a_3', a_6'$ & $a_4', a_7'$\\
    \midrule
    1 & 0.23 & 0.22 & - & 0.24 & 0.16 & 0.11 \\
    2 & 0.23 & 0.22 & - & 0.24 & 0.16 & 0.11 \\
    3 & 0.23 & 0.12 & 0.10 & 0.24 & 0.16 & 0.11 \\
    4 & 0.23 & 0.12 & 0.10 & 0.24 & 0.16 & 0.11 \\
    \bottomrule
    \end{tabular}
\label{tab:4}
\end{table}

This represents an attempt to construct the hand dimensions to be similar to those of a human hand, and different results may be obtained if these values are modified. In addition, the offsets $a_0'$ and $a_1'$, which correspond to the offsets from the palm, were set to 0.1 of the overall hand length. Also, the length of fingers and thumb were set to 0.45 and 0.51 of the overall hand length. For the three DoF finger in Case 1, there is one fewer F/E joint. In this case, the length of the third segment was included in the length of the second segment.

It should be noted that the dimensions presented in Section 3 are not absolute values. The kinematic parameters should be appropriately adjusted according to the user's design intent, mechanism, and mechanical components arrangement. The kinematic structures considered in this work were implemented through adjustments based on existing studies. Even when different configurations are used, it is necessary to modify and apply the parameters appropriately based on the structural characteristics.

%%%%%%%%%%%%%%%%%%%%%%%%%%%%%%%%%%%%%%%%%%%%%%%%%%%%%%%%%%%%%%%%%%%%%%
\section{Evaluation Methods}

Pinch configuration refers to the geometric relationship between the thumb and the opposing finger during pinch contact. To evaluate different pinch configurations, the following metrics are considered. First, a method for evaluating pulp pinch, jaw chuck pinch, and the pulp pinch region performed without the thumb is introduced based on the distal phalanx alignment between the thumb and the other fingers. Next, a method for evaluating lateral pinch performed by the thumb and the index finger is described. Finally, a method for evaluating tip pinch performed by the thumb and the remaining fingers is presented.

\subsection{Distal phalanx alignment}

\begin{figure}[!t]
    \centering
    \includegraphics[width=0.9\columnwidth]{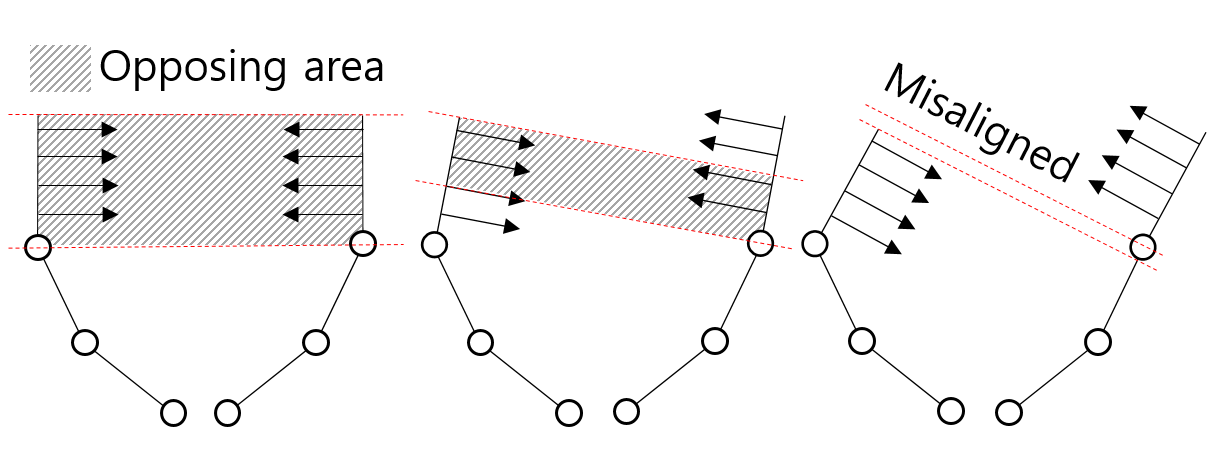}
    \caption{Opposing area based on two-finger facing configuration}
    \label{fig:2}
\end{figure}

Figure ~\ref{fig:2} illustrates the opposing area between two fingers according to their relative postures. Even when point contact occurs between the fingertip and the object during grasping, an opposing area must exist to ensure a stable grasp using the fingertips.

\begin{figure}[!t]
    \centering
    \includegraphics[width=0.9\columnwidth]{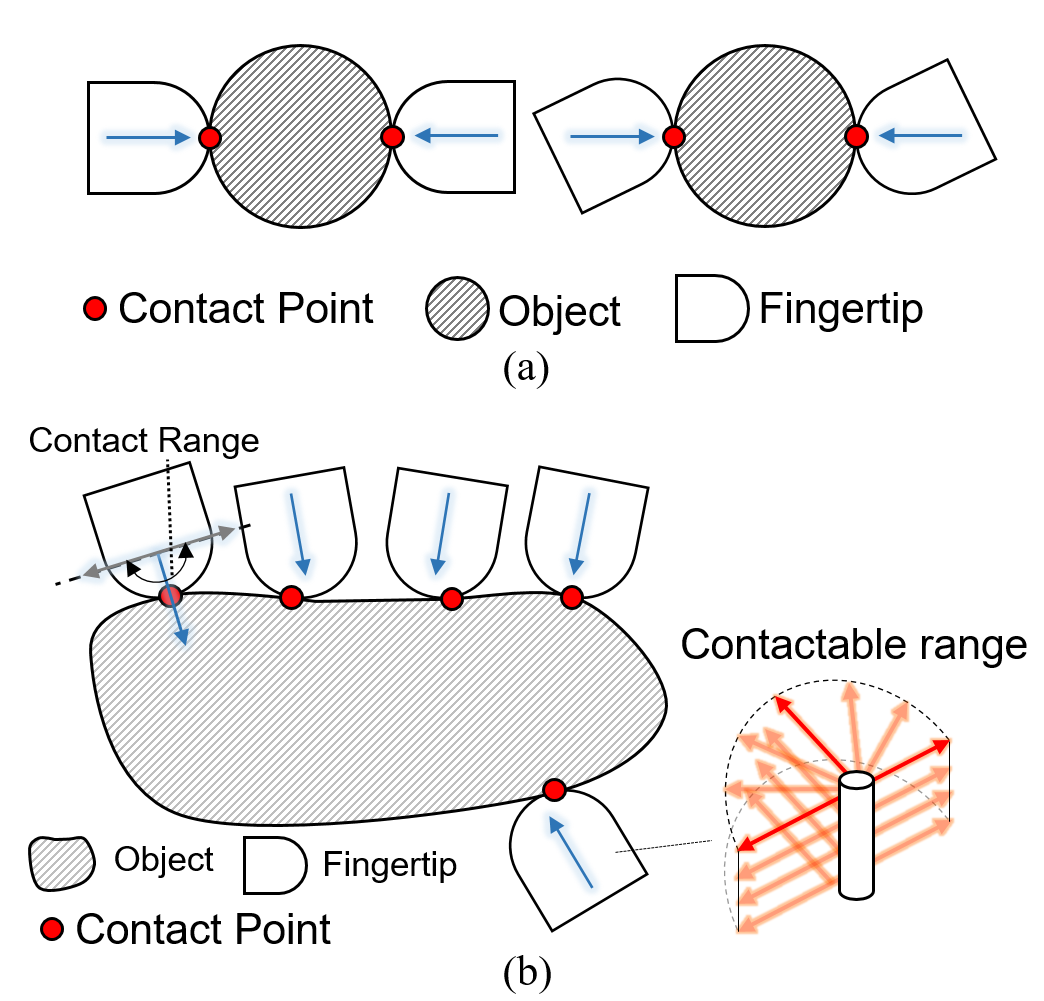}
    \caption{Contactable range avilable for grasping when the fingertip contact surface is assumed as a cylinder}
    \label{fig:3}
\end{figure}

To evaluate such conditions, the normal vector of the fingertip directed toward the object can be utilized. Factors such as finger thickness are not considered. However, in a real human hand, the curved surface of the fingertip enables stable grasping across various finger postures. Figure ~\ref{fig:3}(a) presents an example of this situation. To evaluate grasp conditions regardless of finger posture while considering the possible directions of the fingertip, the phalanx containing the fingertip is modeled as a cylinder. The normal vectors of each finger, denoted as $n_i$ ($i=1$ to $5$ ), are shown in Fig. ~\ref{fig:1}.

\begin{figure}[!t]
    \centering
    \includegraphics[width=0.9\columnwidth]{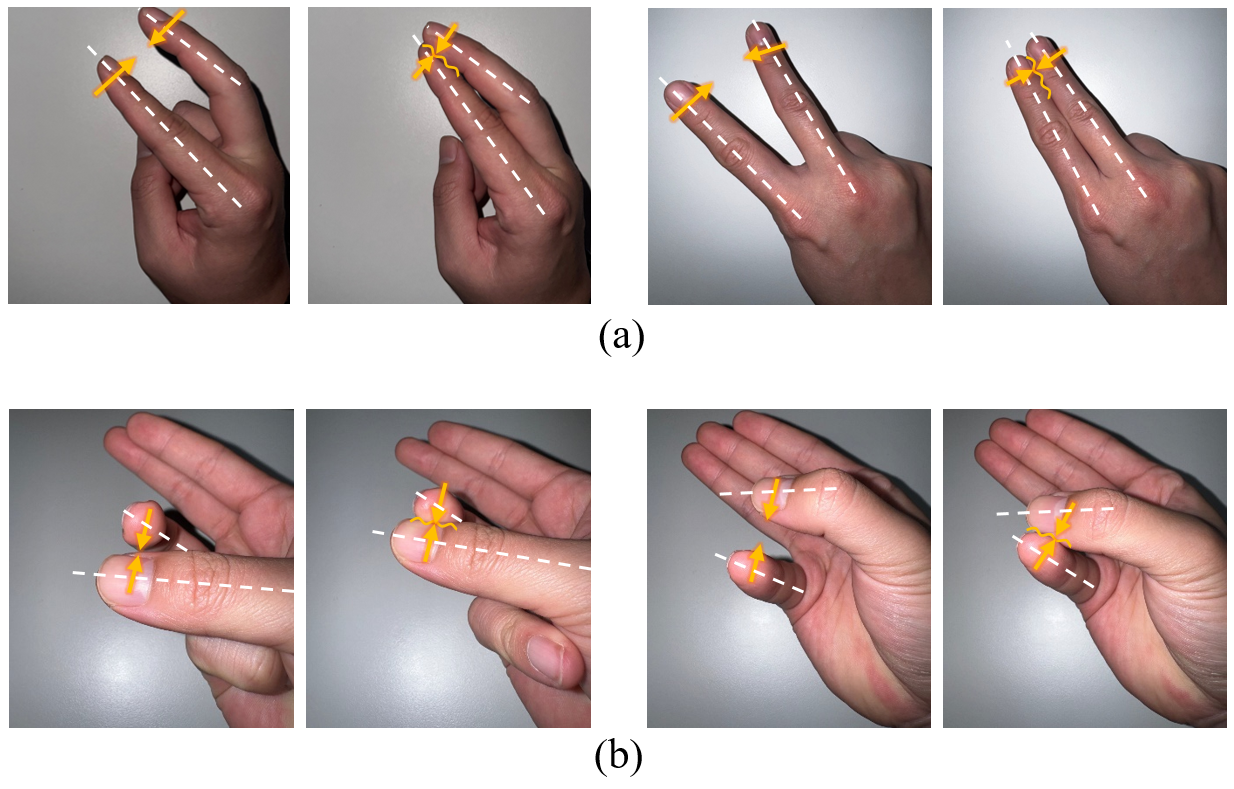}
    \caption{Examples of the contactable state: (a) without using the thumb, (b) using the thumb}
    \label{fig:4}
\end{figure}

Figure ~\ref{fig:3}(b) shows a cross-sectional contact configuration when an object is grasped using five fingertips, where the height dimension is not considered. Since the directions of the fingertip normal vectors differ from one another, the contactable range of each fingertip is illustrated on the right side of Fig ~\ref{fig:3}(b). In practice, the normal vector directions of the fingertips can vary over a wide range. This wide range is assumed to account for grasping configurations such as those shown in Fig. ~\ref{fig:4}(b), where the object may be grasped using the thumb and little finger. Accordingly, the feasible contact region for each fingertip is defined within the range of -90$^\circ$ to 90$^\circ$, as shown in Fig. ~\ref{fig:3}(b). However, without thumb involvement, the contact configuration between the index finger and middle finger can vary, including cases such as those shown in Fig ~\ref{fig:4}(a), where the index finger is above the middle finger, the index finger is below the middle finger, or the middle finger contacts the radial side of the index finger. Therefore, the contactable range was exceptionally set to allow up to 360$^\circ$.

For pinch configurations involving the thumb, the evaluation assumes that pinch grasping is performed based on the opposability of the thumb when the pinch configuration of the five fingertips is considered. This assumption reflects the characteristic motion of the thumb, which enables grasping by opposing the other fingers. The vectors pointing from the distal joint to the fingertip should be mutually parallel at the final rotational joint of each finger. Algorithm ~\ref{alg:1} illustrates the evaluation procedure for distal phalanx alignment.

\begin{algorithm}
\caption{Distal Phalanx Alignment Evaluation}
\label{alg:1}
\begin{algorithmic}[1]
\Require $Q_t, Q_f$ : set of thumb and finger joint configurations
\Ensure $W_t, W_f$ : set of evaluated joint configurations
\State Initialize empty sets $W_t, W_f$ \quad //$t$: thumb, $f$: finger
\For{each thumb configuration $q_t \in Q_t$}
    \State Compute thumb last joint $p_{jt}$ and fingertip $p_{et}$
    \State $\mathrm{v}_t \gets \text{normalize}(p_{et} - p_{jt})$

    \For{each $q_f \in Q_f$ where $f \in \{i,m,r,l\}$}
        \State Compute finger last joint $p_{jf}$ and fingertip $p_{ef}$
        \State $\mathrm{v}_f \gets \text{normalize}(p_{ef} - p_{jf})$

        \If{$|1 - (\mathrm{v}_t \cdot \mathrm{v}_f)| < \epsilon$}  \quad //parallel condition

            \State $\alpha_1 \gets \mathrm{v}_t \cdot p_{jt}, \quad \alpha_2 \gets \mathrm{v}_t \cdot p_{et}$
            \State $\beta_1 \gets \mathrm{v}_t \cdot p_{jf}, \quad \beta_2 \gets \mathrm{v}_t \cdot p_{ef}$

            \State $I_t \gets [\min(\alpha_1,\alpha_2), \max(\alpha_1,\alpha_2)]$
            \State $I_f \gets [\min(\beta_1,\beta_2), \max(\beta_1,\beta_2)]$

            \State $L_{ovr} \gets \min(\max I_t, \max I_f) - \max(\min I_t, \min I_f)$

            \If{$L_{ovr} > 0$}
                \State Add $p_{et}$ to $W_t$
                \State Add $p_{ef}$ to $W_f$
            \EndIf

        \EndIf
    \EndFor
\EndFor
\State \Return $W_t, W_f$
\end{algorithmic}
\end{algorithm}

The distal joint positions and fingertip positions of the thumb and the other fingers are defined. For the thumb, $P_{jt}, P_{et} \in \mathbb{R}^3$ denote the distal joint position and the fingertip position of the thumb, respectively. For the other fingers, $P_{jf}, P_{ef} \in \mathbb{R}^3$ ($f\in{i,m,r,l}$) represent the distal joint position and the fingertip position of each finger, respectively, where $i, m, r,$ and $l$ denote the index, middle, ring, and little fingers.

Each position is calculated using forward kinematics as follows.

\begin{equation}
    P_{et} = FK_t(q_t) \quad\mathrm{and}\quad P_{ef} = FK_f(q_f).
\end{equation}

Here, $q_t$ and $q_f$ denote the joint angles of the thumb and the other fingers, respectively. The unit direction vectors of the distal phalanges are defined as follows:

\begin{equation}
    \mathrm{v}_t = \frac{P_{et}-P_{jt}}{\lVert P_{et}-P_{jt} \rVert} \quad\mathrm{and}\quad \mathrm{v}_f = \frac{P_{ef}-P_{jf}}{\lVert P_{ef}-P_{jf} \rVert}.
\end{equation}

Here, the vectors $\mathrm{v}_t$ and $\mathrm{v}_f$ are normalized vectors with unit magnitude. To satisfy the parallel alignment condition, the distal links of the thumb and the finger must have the same axial direction, which is expressed by the following condition:

\begin{equation}
    \lvert 1 - \mathrm{v}_t^T \mathrm{v}_f \rvert < \epsilon.
\end{equation}

where $\epsilon$ denotes a numerical tolerance and indicates the case in which the two vectors have nearly identical directions. This condition verifies axial alignment.

However, the parallel condition alone is insufficient to determine whether the two vectors are located in positions where interaction is possible. Therefore, as illustrated in Fig. ~\ref{fig:2}, an overlapping region between the opposing finger segments should exist. The investigation of the overlapping region can be divided into two cases: when the finger overlaps the thumb from the upper side and when it overlaps from the lower side.

To evaluate whether the thumb and the opposing finger form a feasible configuration, the distal phalanges are modeled as line segments rather than being described only by directional relationships. Specifically, the distal segments of the thumb and the finger are defined as

\begin{subequations}
\begin{align}
    S_t = \{P_{jt} + \lambda(P_{et}-P_{jt}) \quad | \quad 0 \leq \lambda \leq 1 \},\\
    S_f = \{ P_{jf} + \mu (P_{ef} - P_{jf}) \quad | \quad 0 \leq \mu \leq 1 \}.
\end{align}
\end{subequations}

where $P_{jt}$ and $P_{et}$ denote the base and tip of the thumb distal phalanx, and $P_{jf}$ and $P_{ef}$ denote the middle and distal endpoints of the opposing finger. Overlap is evaluated along the axial direction of the thumb distal segment. The projected scalar coordinates of the segment endpoints are defined as

\begin{subequations}
    \begin{align}
        \alpha_1 = \mathrm{v}_t^T P_{jt}, \alpha_2 = \mathrm{v}_t^T P_{et},\\
        \beta_1 = \mathrm{v}_t^T P_{jf}, \beta_2 = \mathrm{v}_t^T P_{ef}.
    \end{align}
\end{subequations}

The projected intervals of the thumb and the finger distal segments are expressed as

\begin{subequations}
    \begin{align}
        I_t = [ \min(\alpha_1,\alpha_2),max(\alpha_1,\alpha_2) ],\\
        I_f = [ \min(\beta_1,\beta_2), \max(\beta_1,\beta_2) ].
    \end{align}
\end{subequations}

The overlap length $L_{ovr}$ between the two projected intervals is then given by

\begin{equation}
    L_{ovr} = \min(\max I_t, \max I_f) - \max(\min I_t, \min I_f).
\end{equation}

A valid configuration is finally identified when satisfying $L_{ovr} \geq 0$. This ensures that the thumb and finger distal phalanges are not only aligned in direction, but also geometrically share a common axial region, thereby providing a stricter and physically more meaningful criterion for pinch feasibility.

Next, for the case where the thumb is not used, the objective is to determine the region that has a common contactable range among the four remaining fingertips excluding the thumb. Similarly, the condition that the fingertip $x$-axis direction vectors are parallel is applied, and the overlap condition between finger segments can be obtained by changing the reference from the thumb to the index finger. By replacing the thumb with the index finger in Algorithm ~\ref{alg:1} and defining the middle to little fingers as $Q_f$, the corresponding results can be obtained.

\subsection{Lateral pinch}

Lateral pinch is defined as a grasping motion in which an object is grasped using the distal segment of the thumb and all phalanges of the index finger. Algorithm ~\ref{alg:2} presents the lateral pinch evaluation procedure. First, the contactable region of the index finger is defined as the half of the contactable area in Fig. ~\ref{fig:3} that faces the thumb. Here, the contactable points according to the joint configurations of the index finger are defined along the straight line segments connecting adjacent joints.

\begin{algorithm}[!t]
\caption{Lateral Pinch Evaluation}
\label{alg:2}
\begin{algorithmic}[1]
\Require $Q_t, Q_i$ : set of thumb and index joint configurations
\Ensure $W_t, W_i$ : set of evaluated configurations
\State Initialize empty sets $W_t, W_i$ \quad //$t$: thumb, $i$: index finger
\For{each $q_t \in Q_t$}
    \State Compute thumb joint position $p_{jt}$ and fingertip $p_{et}$
    \For{each $q_i \in Q_i$}
        \State Compute index joint position $p_{ji}$ and fingertip $p_{ei}$
        \For{each $d_{span} \in D_{span}$}
            \For{$s_t = 0$ to $1$ step $\Delta s_1$}
                \State $p_{lat} \gets p_{jt} + s_t (p_{et} - p_{jt})$ 
                \quad //on thumb
                \For{$s_i = 0$ to $1$ step $\Delta s_1$}
                    \State $p_{lai} \gets p_{ji} + s_i (p_{ei} - p_{ji})$ 
                   \quad //on index
                    \If{$\left| \|p_{lat} - p_{lai}\| - d_{span} \right| < \epsilon$}
                        \If{$p_{lat}(y_o) \ge p_{lai}(y_o)$}
                        \State $\mathcal{S}_i \gets$ set of index phalanx
                            \If{$p_{lat}(x_o) \in \mathcal{S}_i$}
                                \State Add $p_{et}$ to $W_t$
                                \State Add $p_{ei}$ to $W_i$
                            \EndIf
                        \EndIf
                    \EndIf
                \EndFor
            \EndFor
        \EndFor
    \EndFor
\EndFor
\State \Return $W_t, W_i$
\end{algorithmic}
\end{algorithm}

As described in Section 4.1, the distance between the unit direction vector $\mathrm{v}_t$ of the distal phalanx of the thumb and the region formed by the joints of the index finger is denoted as $d_{lp}$. Let the maximum span of the lateral pinch be $d_{lpm}$. By considering the joint configurations of the thumb and index finger and the distance between the corresponding regions as variables, the feasible region for lateral pinch can be obtained through the overlap detection described in Section 4.1. Considering the postures of the thumb and index finger, the $y_o$ coordinate of the thumb contact point observed in the coordinate frame $P_o x_o y_o z_o$, as shown in Fig. ~\ref{fig:1}(a), is constrained not to be smaller than the $y_o$ coordinate of the index finger contact point. In addition, the $x_o$ coordinate of the thumb contact point, expressed in the coordinate frame $P_o x_o y_o z_o$, is constrained to lie within the $x_o$-range defined by the corresponding phalanx segment of the index finger. In this process, fingertip points are not detected directly; instead, points within the detected phalanges of each finger are obtained.

\subsection{Tip pinch}

Tip pinch is defined as a grasping motion in which the thumb and the other fingers grasp an object using the fingertip points. Algorithm ~\ref{alg:3} presents the tip pinch evaluation method.

For fingertip grasping, unlike the cylindrical grasp considered in Fig. ~\ref{fig:3}, the evaluation focuses on the finger posture and the distance between the fingertips. Since the finger posture involved in tip pinch corresponds to a configuration where the fingers are inclined toward the palm rather than satisfying the parallel condition described in Section 4.1, the unit direction vector $\mathrm{v}_t$ described in Section 4.1, the unit direction vector $\mathrm{v}_f$ of the distal phalanx of the other finger should not be parallel. Therefore, the following condition should be satisfied:

\begin{equation}
    \lvert \mathrm{v}_t^T \mathrm{v}_f \rvert < 1
\end{equation}

The distal joints of the two fingers should not intersect. Considering the posture between the two fingers, the $z_o$ coordinate of the thumb contact point observed in the coordinate frame $P_o x_o y_o z_o$ shown in Fig. ~\ref{fig:1}(a) is constrained not to be smaller than the $z_o$ coordinate of the contact point of the other finger. As described in Section 4.1, the distance between the thumb endpoint $P_{et}$ and the endpoint of the other finger $P_{ef}$ is denoted as $d_{tp}$, and the maximum span of the tip pinch is denoted as $d_{tpm}$. By considering the joint configurations of the fingers and the distance between the two fingertip points as variables, the region where tip pinch is feasible can be detected. In the tip pinch evaluation, the endpoints of each finger are detected. Also, considering the contactable range of the fingers, in the $P_o x_o y_o z_o$ frame the $y$-axis value of the thumb was constrained to be not smaller than that of the index finger, and the $z$-axis value of the thumb was constrained to be not greater than that of the index finger. 

\begin{algorithm} 
\caption{Tip Pinch Evaluation}
\label{alg:3}
\begin{algorithmic}[1]
\Require $Q_t, Q_f$ : set of thumb and finger joint configurations ($f \in \{i,m,r,l\}$)
\Ensure $W_t, W_f$ : set of evaluated joint configurations
\State Initialize empty sets $W_t, W_f$
\For{each $q_t \in Q_t$}
    \State Compute thumb last joint $p_{jt}$ and fingertip $p_{et}$
    \State $\mathrm{v}_t \gets \text{normalize}(p_{et} - p_{jt})$
    \For{each $q_f \in Q_f$}
        \State Compute finger last joint $p_{jf}$ and fingertip $p_{ef}$
        \State $\mathrm{v}_f \gets \text{normalize}(p_{ef} - p_{jf})$
        \State $d_f \gets \|p_{et} - p_{ef}\|$ //distance between fingertips
        \For{each $d_{span} \in D_{span}$}
            \If{$1 - |\mathrm{v}_t \cdot \mathrm{v}_f| > 0$}
            //non-parallel
                \If{$p_{et}(y_o) \ge p_{ef}(y_o)$}
                //posture
                    \If{$p_{ef}(z_o) \ge p_{et}(z_o)$}
                    //posture
                        \If{$|d_f - d_{span}| < \epsilon$}
                            \State Add $p_{et}$ to $W_t$
                            \State Add $p_{ef}$ to $W_f$
                        \EndIf
                    \EndIf
                \EndIf
            \EndIf
        \EndFor
    \EndFor
\EndFor
\State \Return $W_t, W_f$
\end{algorithmic}
\end{algorithm}

Based on the three methods described above, this study can detect the pinch configurations that can be generated by a five-fingered hand. By classifying and evaluating each pinch motion, the results can be used to estimate the performance of the kinematic structure during the design process.

%%%%%%%%%%%%%%%%%%%%%%%%%%%%%%%%%%%%%%%%%%%%%%%%%%%%%%%%%%%%%%%%%%%%%%
\section{Analysis and Results}
In this section, the process and results of deriving the detected points relative to the reachable points are described according to the four proposed kinematic structure configurations and the pinch motions. First, forward kinematics is performed, and based on this, the reachable region is derived according to the allowable joint angle ranges of each joint. Subsequently, the results are obtained using the detection algorithms described in the previous section. For the first case, distal phalanx alignment is described for all four cases, and in this process, the results according to the tolerance and the detection angle interval are presented together. For lateral pinch and tip pinch, the results of the four cases are presented according to the specified tolerance, detection angle interval, and detection distance.

\begin{table}[!h]
    \caption{DH Table for five DoF Thumb}
    \centering
    \begin{tabular}{c|c|c|c|c}
    \toprule
    $i$ & $\alpha_{i-1}$ & $a_{i-1}$ & $d_i$ & $\theta_i$ \\
    \midrule
    1 & 0 & $a_0'$ & 0 & $\theta_1'$ \\
    2 & -$\pi$/2 & $a_1'$ & 0 & $\theta_2'$ \\
    3 & $\pi$/2 & $a_2'$ & 0 & $\pi$/2 \\
    4 & $\pi$/2 & 0 & 0 & $\theta_4'$ \\
    5 & -$\pi$/2 & 0 & 0 & -$\pi$/2 \\
    6 & -$\pi$/2 & 0 & 0 & $\theta_6'$ \\
    7 & 0 & $a_6'$ & 0 & $\theta_7'$ \\
    8 & 0 & $a_7'$ & 0 & 0 \\
    \bottomrule
    \end{tabular}
\label{tab:5}
\end{table}

\begin{table}[!h]
    \caption{DH Table for four DoF Middle Finger}
    \centering
    \begin{tabular}{c|c|c|c|c}
    \toprule
    $i$ & $\alpha_{i-1}$ & $a_{i-1}$ & $d_i$ & $\theta_i$ \\
    \midrule
    1 & $\pi$/2 & 0 & 0 & $\pi$/2 \\
    2 & $\pi$/2 & $a_1$ & $d_2$ & $\theta_2$ \\
    3 & -$\pi$/2 & 0 & 0 & $\theta_3$ \\
    4 & 0 & $a_3$ & 0 & $\theta_4$ \\ 
    5 & 0 & $a_4$ & 0 & $\theta_5$ \\
    6 & 0 & $a_5$ & 0 & 0 \\
    \bottomrule
    \end{tabular}
    \label{tab:6}
\end{table}

\begin{table}[!h]
    \caption{DH Table for four DoF Thumb}
    \centering
    \begin{tabular}{c|c|c|c|c}
    \toprule
    $i$ & $\alpha_{i-1}$ & $a_{i-1}$ & $d_i$ & $\theta_i$ \\
    \midrule
    1 & 0 & $a_0'$ & 0 & $\theta_1'$ \\
    2 & -$\pi$/2 & $a_1'$ & 0 & $\theta_2'$ \\
    3 & $\pi$/2 & $a_2'$ & 0 & $\theta_3'$ \\
    4 & 0 & $a_3'$ & 0 & $\theta_4'$ \\
    5 & 0 & $a_4'$ & 0 & 0 \\
    \bottomrule
    \end{tabular}
    \label{tab:7}
\end{table}

\begin{table}[!h]
    \caption{DH Table for three DoF Middle Finger}
    \centering
    \begin{tabular}{c|c|c|c|c}
    \toprule
    $i$ & $\alpha_{i-1}$ & $a_{i-1}$ & $d_i$ & $\theta_i$ \\
    \midrule
    1 & $\pi$/2 & 0 & 0 & $\pi$/2 \\
    2 & $\pi$/2 & $a_1$ & $d_2$ & $\theta_2$ \\
    3 & -$\pi$/2 & 0 & 0 & $\theta_3$ \\
    4 & 0 & $a_3$ & 0 & $\theta_4$ \\
    5 & 0 & $a_4$ & 0 & 0 \\
    \bottomrule
    \end{tabular}
    \label{tab:8}
\end{table}

\subsection{Forward kinematics and reachable area}

To consider all reachable regions, forward kinematics must be performed. Let $T_i^j$ denote the homogeneous transformation matrix from frame $i$ to $j$ (where $j > i$). This matrix can be obtained through the following sequence of transformations:

\begin{equation}
    T_i^j = T_i^{i+1} T_{i+1}^{i+2}\cdots T_{j-2}^{j-1} T_{j-1}^{j}.
\end{equation}

First, the case with the largest number of DoF, Case 4 (Table ~\ref{tab:4}), is considered. Each transformation matrix can be derived using the DH parameter tables in Tables ~\ref{tab:5} and ~\ref{tab:6}. Table ~\ref{tab:5} represents the five DoF thumb, and Table ~\ref{tab:6} represents the four DoF middle finger. For the other fingers except the middle finger, the representation can be obtained by modifying the offset $a_{i-1}$ in column $i=2$ of the DH table. Using the DH parameter convention, the joints of the fingers are represented as interconnected through a kinematic chain of links \cite{c31}. By multiplying each transformation matrix, the transformation from initial to the end frame can be obtained. 

Cases 1, 2, and 3 indicated in Table ~\ref{tab:4} can be derived by modifying the DH tables in Tables ~\ref{tab:5}, ~\ref{tab:6}, ~\ref{tab:7}, and ~\ref{tab:8}. The actual values of the parameters are expressed as ratios using Tables ~\ref{tab:3} and ~\ref{tab:4}, where the total hand length is normalized to 1. Based on this, the joint positions and end positions for each of the four kinematic structures can be represented with respect to the coordinate frame $P_o x_o y_o z_o$.

\begin{figure}[!t]
    \centering
    \includegraphics[width=0.95\columnwidth]{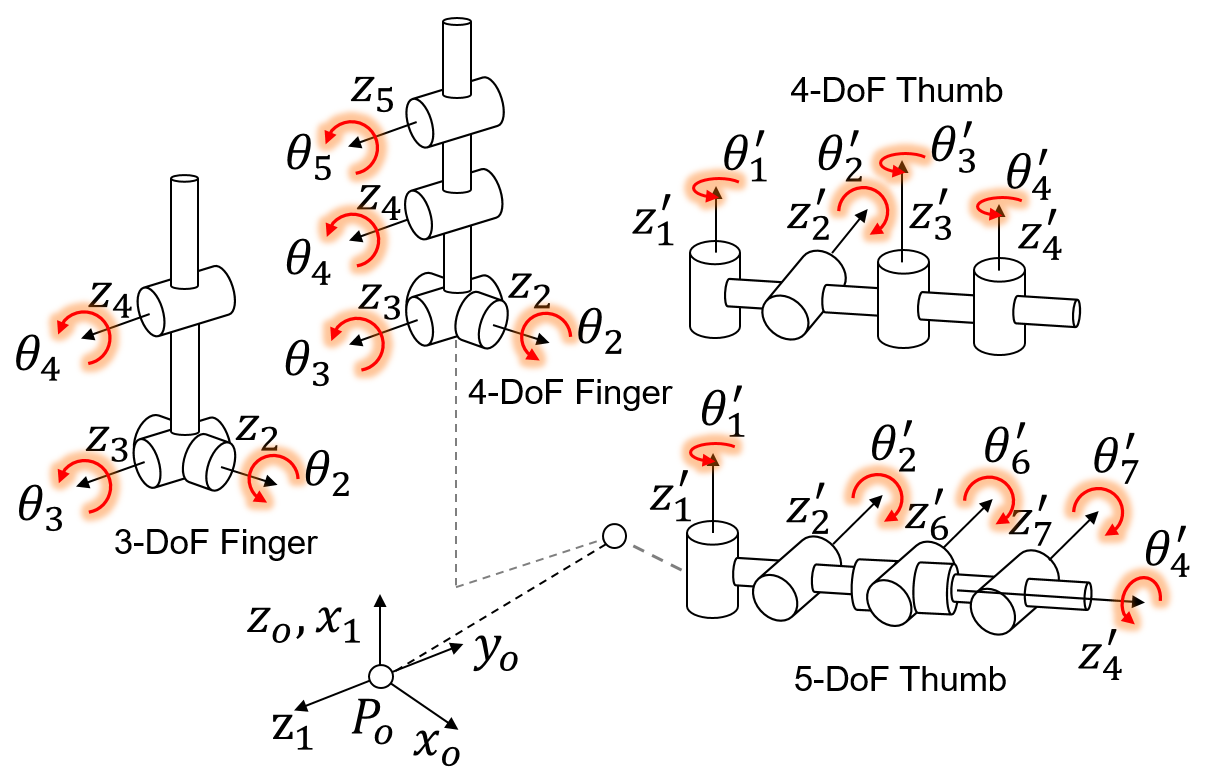}
    \caption{Initial posture of thumb and fingers for four cases}
    \label{fig:5}
\end{figure}

The initial postures of the thumb and the other fingers for the four cases are shown in Fig. ~\ref{fig:5}. The joint motion ranges are defined with respect to the initial posture. The motion ranges of the fingers are listed in Table ~\ref{tab:9}, and those of the thumb are listed in Table ~\ref{tab:10}. For the finger, the actuating range is set differently only for tip pinch in order to verify that the computation properly adapts to changes in the actuating range.

\begin{table}[h]
    \caption{Actuating ranges of fingers for four cases}
    \centering
    \begin{tabular}{c|c|c|c|c}
    \toprule
    Case & $\theta_2$ & $\theta_3$ & $\theta_4$ & $\theta_5$ \\
    \midrule
    1 & -$\pi$/6 to $\pi$/6 & -$\pi$/2 to $2\pi$/9 & -$\pi$/2 to 0 & - \\
    2 & -$\pi$/6 to $\pi$/6 & -$\pi$/2 to $2\pi$/9 & -$\pi$/2 to 0 & -$\pi$/2 to 0 \\
    3 &-$\pi$/6 to $\pi$/6 & -$\pi$/2 to $2\pi$/9 & -$\pi$/2 to 0 & - \\
    4 &-$\pi$/6 to $\pi$/6 & -$\pi$/2 to $2\pi$/9 & -$\pi$/2 to 0 & -$\pi$/2 to 0 \\
    \bottomrule
    \end{tabular}
    \label{tab:9}
\end{table}

\begin{table}[h]
    \caption{Actuating ranges of thumb for four cases}
    \centering
    \begin{tabular}{c|c|c|c|c}
    \toprule
    Case & $\theta_1'$ & $\theta_2'$, $\theta_4'$ & $\theta_3'$ & $\theta_6'$, $\theta_7'$ \\
    \midrule
    1 & 0 to $\pi$/2 & -$\pi$/2 to 0 & -$\pi$/2 to 0  & - \\
    2 & 0 to $\pi$/2 & -$\pi$/2 to 0 & -$\pi$/2 to 0  & - \\
    3 & 0 to $\pi$/2 & -$\pi$/2 to 0 & -$\pi$/6 to $\pi$/6 & -$\pi$/2 to 0  \\
    4 & 0 to $\pi$/2 & -$\pi$/2 to 0 & -$\pi$/6 to $\pi$/6 & -$\pi$/2 to 0  \\
    \bottomrule
    \end{tabular}
    \label{tab:10}
\end{table}

\begin{figure}[!t]
    \centering
    \includegraphics[width=0.9\columnwidth]{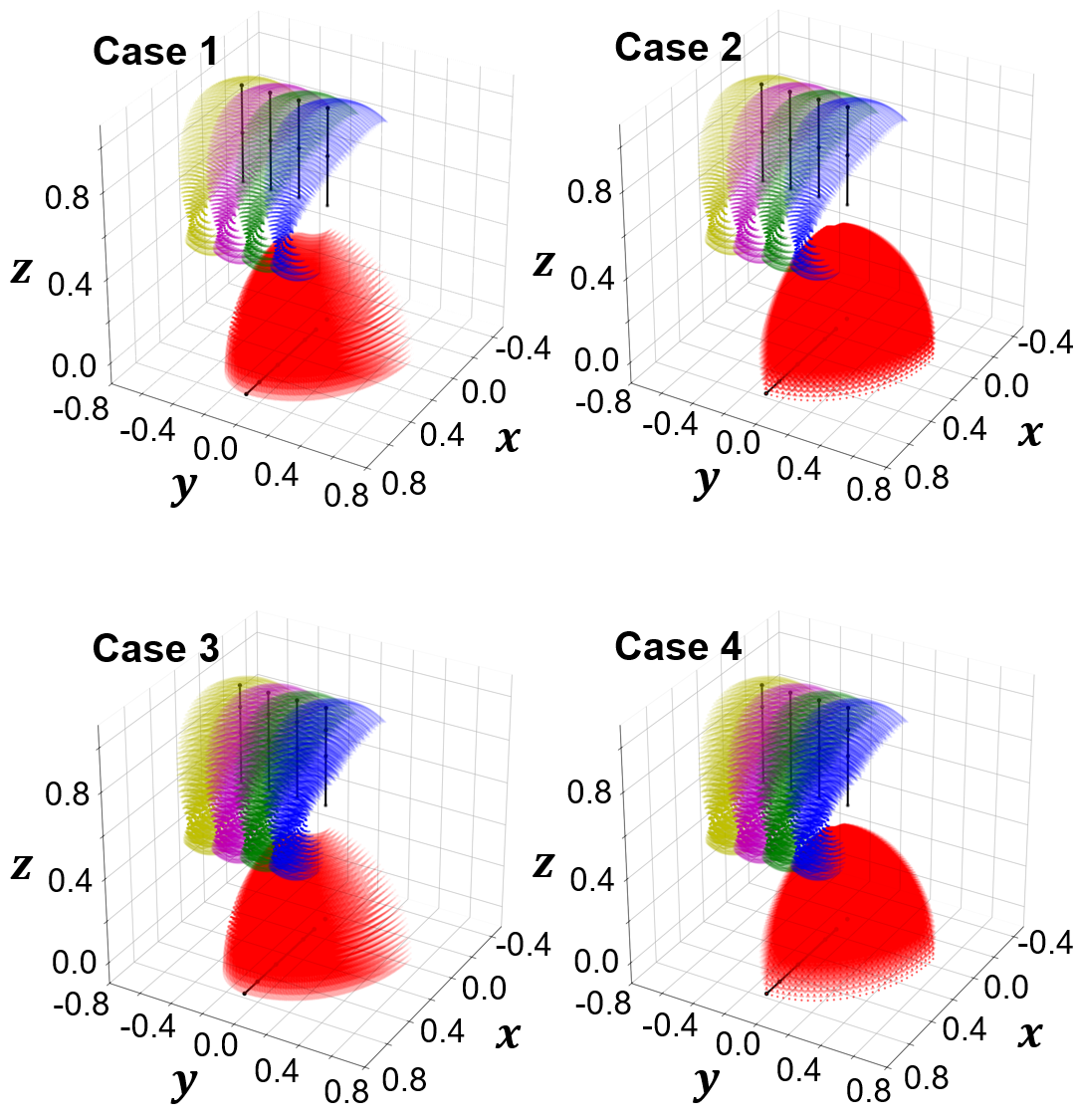}
    \caption{Reachable areas for four cases with different kinematic structure}
    \label{fig:6}
\end{figure}

For each case, the reachable area of the fingertip from the initial posture is shown in Fig. ~\ref{fig:6}. Although the overall hand size in the initial posture is identical for all cases, differences in the reachable areas occur depending on the kinematic structure. In particular, for the fingers, since an additional DoF is added to the F/E motion for the same link length, the reachable area becomes larger. In the case of the thumb, the reachable area differs because the motion characteristics vary depending on the structural configuration.

\subsection{Evaluation of distal phalanx alignment}

Aligned area was evaluated by setting the tolerance for the inner product and vector overlap as $\epsilon$. This tolerance is introduced because performing computations for all reachable points is computationally complex. In particular, to compute the area where the other fingers ($O_f$, $f=i,m,r,l$) can reach with respect to the set of all points reachable by the thumb ($O_t$), a total of $O_t \times O_f$ point-wise operations are required. This area varies depending on the number of steps used to discretize the joint motion ranges.

\begin{figure}[!t]
    \centering
    \includegraphics[width=0.85\columnwidth]{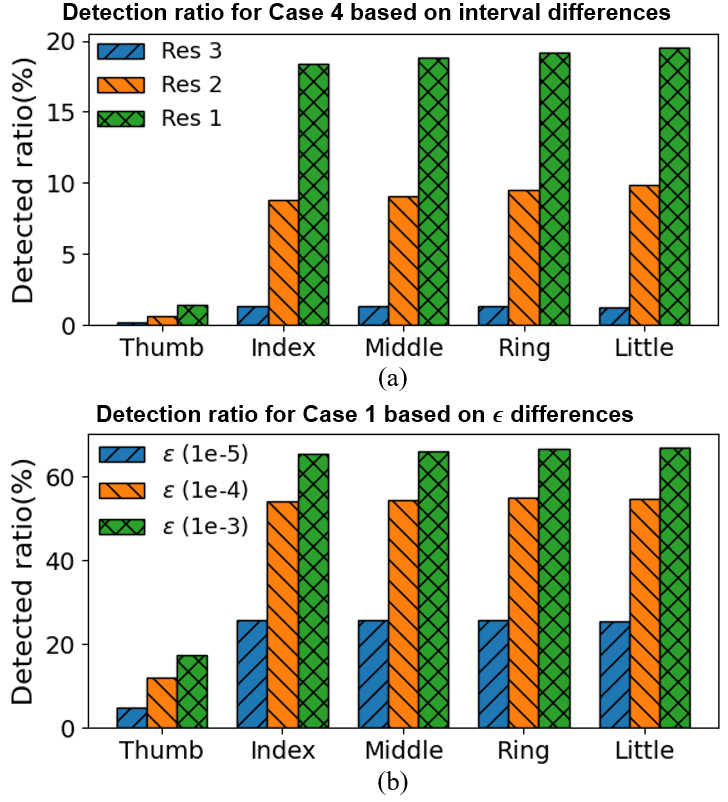}
    \caption{Detection ratio of the distal phalanx aligned area within the reachable area (a) according to the resolution, (b) according to the $\epsilon$}
    \label{fig:8}
\end{figure}

First, to examine the results according to the number of steps, $\epsilon=1\times10^{-5}$ was used, and the joint angles were discretized into three resolution condition. Three resolution conditions were applied. For condition 1 (Res 1), the actuating range was divided into 9 intervals for 90$^\circ$, 6 intervals for 60$^\circ$, and 11 intervals for 130$^\circ$. For condition 2 (Res 2), the actuating range was divided into 18 intervals for 90$^\circ$, 12 intervals for 60$^\circ$, and 22 intervals for 130$^\circ$. For condition 3 (Res 3), the actuating range was divided into 30 intervals for 90$^\circ$, 20 intervals for 60$^\circ$, and 33 intervals for 130$^\circ$. The corresponding results are presented in Fig. ~\ref{fig:8}(a). Fig.~\ref{fig:8}(a) shows the results for Case 4, which requires the largest number of computations.

Within the same kinematic structure, when the joint angles are discretized with larger intervals, the number of detection steps decreases, resulting in fewer postures that can be evaluated. Consequently, the number of points detected as the dexterity area also decreases. Fig. ~\ref{fig:8}(a) shows, for each finger, the number of detected points relative to the total number of evaluated points under the three resolution conditions.

As the interval is reduced and the detection becomes denser, the proportion of detected points within the reachable area increases for all fingers. This is due to the increased computational cost, which increases depending on the number of finger DoF and the interval used for the joint angle discretization. When the interval is too large, it becomes difficult to obtain appropriate results. 

Next, it is necessary to determine an appropriate value of $\epsilon$. When the resolution is increased and more points are detected, regions can still be detected even with smaller values of $\epsilon$. However, since the detection is not performed for the entire continuous area, there is a high possibility that some areas may not be detected depending on the resolution scale.

Figure ~\ref{fig:8}(b) shows the detected areas obtained with Res 3 while varying the value of $\epsilon$ as $1\times 10^{-3}$, $1\times 10^{-4}$, and $1\times 10^{-5}$. Compared to the same evaluated area, the detected area decreases as the value of $\epsilon$ approaches zero. For the thumb, the detected area relative to the total area decreases from 17.43\% to 12.01\% and 4.95\%. The four fingers show the same tendency, and the average detection ratios of the four fingers decrease from 66.28\% to 54.61\% and 25.68\%.

\begin{figure}[!t]
    \centering
    \includegraphics[width=0.95\columnwidth]{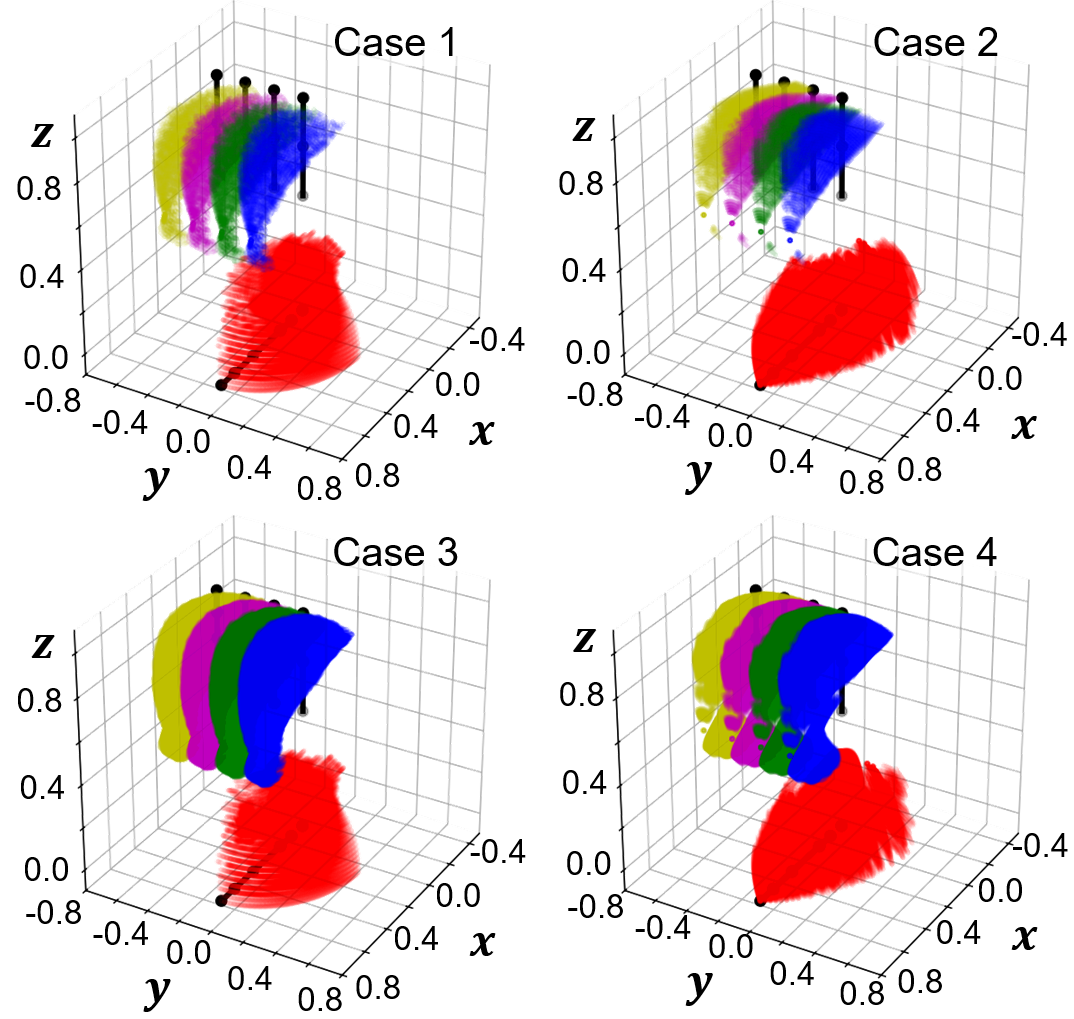}
    \caption{Distal phalanx alignment detection results for each case with $\epsilon = 1 \times 10^{-5}$, Res 3}
    \label{fig:10}
\end{figure}

Next, under the same interval and $\epsilon$, the detection results for each case are shown in Fig. ~\ref{fig:10}. The number of evaluated points in the inspection area for each finger according to the interval for each case is presented in Table ~\ref{tab:9}, and the number of detected points according to the value of $\epsilon$ is presented in Table ~\ref{tab:10}. The number of points in the inspection area tends to increase as the number of DoF increases, since a larger number of configurations must be examined.

\begin{table}[h]
    \caption{Number of evaluated points in the inspection region of each finger for each case ($\epsilon = 1 \times 10^{-5}$, Res 3)}
    \centering
    \begin{tabular}{c|c|c}
    \toprule
    Case & Thumb & Index to Little \\
    \midrule
    1 & 810,000 & 19,800\\
    2 & 16,200,000 & 19,800 \\
    3 & 810,000 & 594,000 \\
    4 & 16,200,000 & 594,000 \\
    \bottomrule
    \end{tabular}
    \label{tab:11}
\end{table}

\begin{table}[h]
    \caption{Number of detected points in the inspection region of each finger for each case ($\epsilon = 1 \times 10^{-5}$, Res 3)}
    \centering
    \begin{tabular}{c|c|c|c|c|c}
    \toprule
    Case & Thumb & Index & Middle & Ring & Little \\
    \midrule
    1 & 38,708 & 4,698 & 4,745 & 4,711 & 4,615 \\
    2 & 115,644 & 3,638 & 3,761 & 3,928 & 4,031 \\
    3 & 38,797 & 67,367 & 70,104 & 73,556 & 76,568\\
    4 & 159,525 & 41,249 & 44,811 & 49,363 & 53,025 \\
    \bottomrule
    \end{tabular}
    \label{tab:12}
\end{table}

An important point is observed when comparing Case 1 with Case 3 and Case 2 with Case 4. In each pair, the thumb configuration is identical, while the DoF of the fingers differ. In this case, when the fingers have three degrees of freedom, increasing the thumb’s degrees of freedom does not significantly improve the detection ratio. In contrast, when the fingers have four degrees of freedom, the detection ratio increases substantially as the thumb’s degrees of freedom increase. This is because the finger configurations are almost identical, except that one additional DoF is added in the F/E motion.

When comparing Case 1 with Case 2 and Case 3 with Case 4, the finger DoF are identical in each pair, while the thumb DoF differ. In this case, not only the number of DoF but also the implemented motions are different. To generate the thumb motion of Case 3 using the configurations of Case 4, the angle $\theta_4'$ must be $\pi/2$ at the initial posture. However, the current motion range of this joint is limited to $-\pi/6$ to $\pi/6$, and therefore Cases 3 and 4 exhibit different motion ranges. If Case 4 had a motion range that included the motion of Case 3, it would produce results over a wider range.

Next, the case without thumb involvement is briefly examined. Similarly, $\epsilon = 1 \times 10^{-5}$, Res 3,and the same detection algorithm is used, but the index finger is used as the reference instead of the thumb to detect the other fingers.

\begin{table}[h]
    \caption{Number of detected points in the inspection region of each finger for each case without thumb ($\epsilon = 1 \times 10^{-5}$, Res 3)}
    \centering
    \begin{tabular}{c|c|c|c|c}
    \toprule
    Case & Index & Middle & Ring & Little \\
    \midrule
    1,2 & 19,800 & 19,800 & 19,800 & 19,115 \\
    3,4 & 594,000 & 594,000 & 572,461 & 526,435  \\
    \bottomrule
    \end{tabular}
    \label{tab:13}
\end{table}

Table ~\ref{tab:13} shows the numbers of detected points while evaluated points are same with Table ~\ref{tab:11}. First, the index finger and the middle finger are detected in both cases. This is because the two fingers have identical DoF configurations and are positioned with a narrow installation spacing. For the ring finger, the results are identical for the three DoF finger, while the detected area decreases only in the four DoF configuration. In the case of the little finger, the detected area decreases in both cases. In both configurations, the length of the distal phalanx in the four DoF case is shorter than that in the three DoF case. As a result, when the fingers face each other due to the A/A motion, there are more cases where the segments do not overlap. In contrast, in the three DoF case, since the distal segment is longer than that of the four DoF case, fewer configurations deviate due to the A/A motion.

\subsection{Evaluation of lateral pinch}

For the lateral pinch evaluation, the resolution condition was set to Res 1 and $\epsilon = 1 \times 10^{-5}$, considering the increase in computational cost caused by the detection according to the span between pinch postures and the distinction of contact points. The joints corresponding to the contact between the index finger and the thumb were divided into ten segments, resulting in eleven contact points. The distance was detected in eleven steps from 0 to 1.0 with an interval of 0.1 based on the total hand length.

\begin{figure}[!t]
    \centering
    \includegraphics[width=0.95\columnwidth]{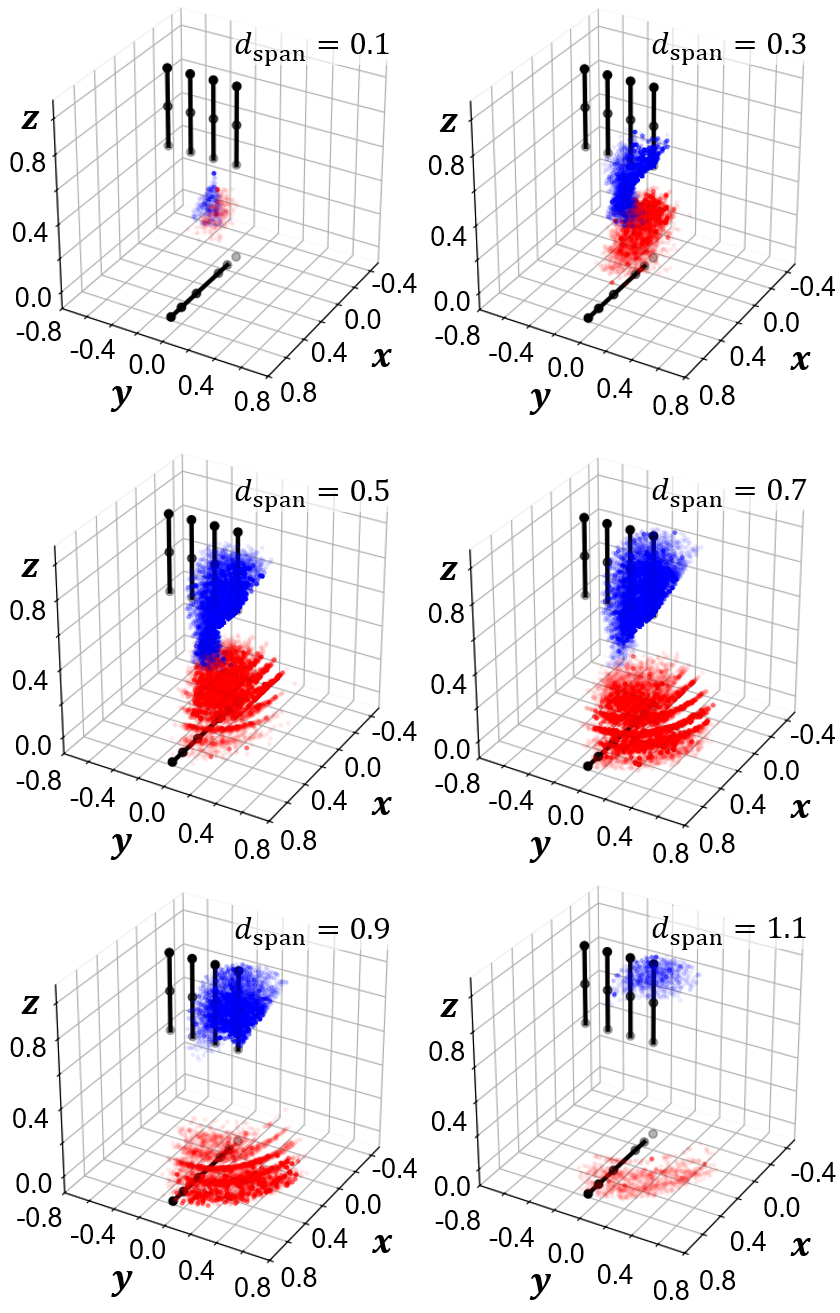}
    \caption{Lateral pinch detection results for Case 1 with $\epsilon = 1 \times 10^{-5}$, Res 1}
    \label{fig:11}
\end{figure}

Figure ~\ref{fig:11} shows the detection results of lateral pinch for case 1. From the graphs obtained for each distance, it can be seen that the degree of detection varies depending on the distance. In particular, the highest number of detections occurred when the distance between the index finger and thumb fingertips, $d_{span}$, was between 0.5 and 0.7, and some postures corresponding to distances greater than 1, which approximately corresponds to the hand length, were also detected. However, when $d_{span}=0$, no corresponding points were obtained due to the small $\epsilon$ and detection interval.

The postures reachable by the thumb include all corresponding configurations with respect to the index finger. The number of detected points for the three DoF finger is 594, while that for the 4 DoF finger is 5,346. The number of detected points for the four DoF thumb is 6,561, and that for the five DoF thumb is 39,366. The number of detected points relative to the evaluated points for each case is presented in Table ~\ref{tab:14}.

\begin{table}[h]
    \caption{Lateral pinch detection results – detected points relative to the evaluated points}
    \centering
    \begin{tabular}{c|c|c}
    \toprule
    Case & Thumb (\%) & Index (\%)  \\
    \midrule
    1 & 44.25 & 35.77  \\
    2 & 73.87 & 32.26 \\
    3 & 79.45 & 36.14 \\
    4 & 88.57 & 36.24 \\
    \bottomrule
    \end{tabular}
    \label{tab:14}
\end{table}

According to the results, when the same thumb is used, the region where lateral pinch is possible in the thumb becomes larger when the finger has four DoF. However, when the same finger configuration is used, the detection ratio of the four DoF thumb is higher than that of the five DoF thumb. This is because, unlike the fingers where the difference occurs in the number of DoF corresponding to the F/E motion, the thumb has structural differences.

Therefore, in the case of the thumb, a higher number of DoF does not necessarily produce higher detection results without considering the structural configuration. In addition, since the distance between contact points for the detection region was defined based on the total hand length, this ratio is suitable for identifying the general tendency. However, to obtain an accurate detection ratio, the computation can be performed by setting the maximum stroke between the index finger and the thumb that can occur in a posture as the maximum calculation range.

\subsection{Evaluation of tip pinch}

For the tip pinch analysis, the resolution condition was set to Res 1 due to the increased computational cost caused by the distinction between detection and contact points and the number of pinch posture combinations. The $\epsilon$ was set to $1 \times 10^{-5}$. The distance between the thumb tip and the tips of the four fingers was evaluated in 13 steps from 0 to 1.2 with an interval of 0.1, normalized by the total hand length.

\begin{figure}[!t]
    \centering
    \includegraphics[width=0.95\columnwidth]{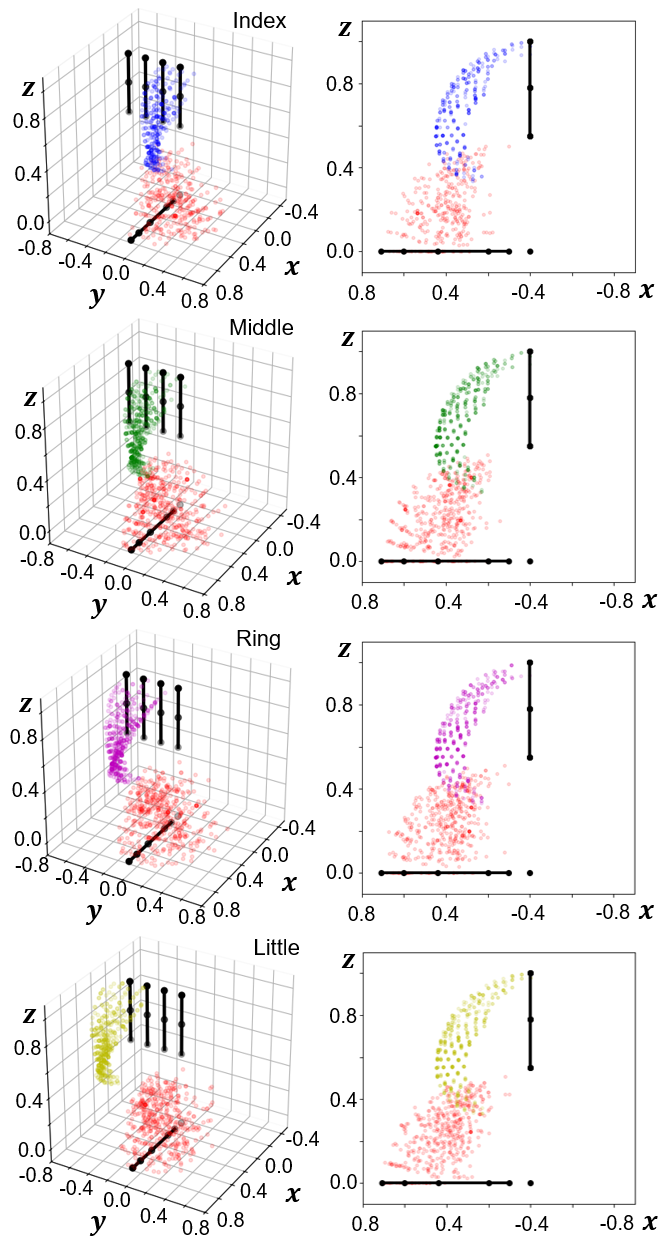}
    \caption{Tip pinch detection results for Case 1 with $\epsilon = 1 \times 10^{-5}$, Res 1.}
    \label{fig:12}
\end{figure}

\begin{figure*}[!t]
    \centering
    \includegraphics[width=0.99\textwidth]{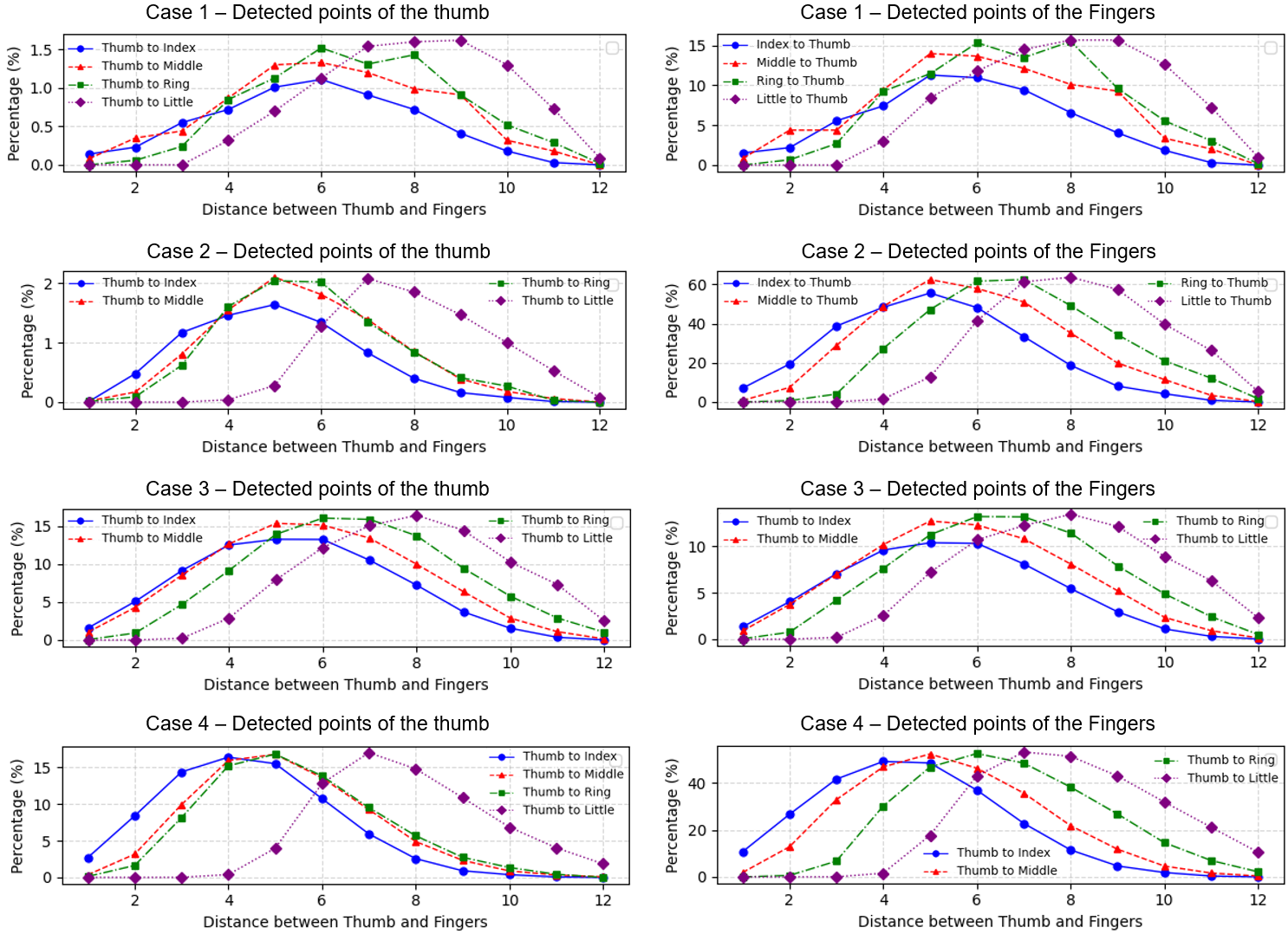}
    \caption{Tip pinch detection results for four cases with $\epsilon = 1 \times 10^{-5}$, Res 1}
    \label{fig:13}
\end{figure*}

Figure ~\ref{fig:12} shows the detection results of tip pinch for case 1. The results are presented by distinguishing between the four fingers and the thumb, while the displayed outcomes do not differentiate based on the distances between the fingertips. The detected points for each finger vary, and excluding duplicates, the total number of detected points is lowest for the index finger, followed by the middle, ring, and little fingers in increasing order. These results allow verification of cases in which a candidate set that can be generally observed is obtainable.

The reachable postures of the thumb include the information of all fingers from the index finger to the little finger. The number of evaluated points is 594 for a three DOF finger and 5,346 for a four DOF finger. For the thumb, the number of evaluated points is 6,561 for the four DOF thumb and 39,366 for the five DOF thumb.

The ratio of detected points to evaluated points for each case is summarized in Table ~\ref{tab:15}. According to the results, when the fingers have the same number of DoF, the five DOF thumb shows a larger detectable region for tip pinch. In addition, when the fingers have four DOF, a larger region capable of performing tip pinch is detected with the thumb.

\begin{table}[h]
    \caption{Tip pinch detection results – detected points relative to the evaluated points}
    \centering
    \begin{tabular}{c|c|c|c|c|c}
    \toprule
    \multirow{2}{*}{Case} & Thumb & Index & Middle & Ring & Little\\
    & (\%) & (\%) & (\%) & (\%) & (\%) \\
    \midrule
    1 & 26.66 & 42.09 & 53.37 & 55.56 & 57.74 \\
    2 & 28.83 & 80.30 & 81.82 & 81.82 & 81.82 \\
    3 & 93.77 & 44.11 & 55.31 & 58.21 & 56.34 \\
    4 & 95.24 & 81.11 & 81.72 & 81.71 & 81.69 \\
    \bottomrule
    \end{tabular}
    \label{tab:15}
\end{table}

For the lateral case, results are presented by distance, but no further distinction was made due to two-finger interaction. For the tip pinch, however, interactions with each finger were evaluated separately, revealing differences depending on the finger combination. When distinguished by finger relationship for each distance, the results are shown in Fig. ~\ref{fig:13}, displaying four different cases. The results indicate the detection ratio per distance based on the reachable area for each interval. Coordinates detected at each distance sometimes occurred multiple times.

In all cases, tip pinch at shorter distances had higher usable ratios for fingers closer to the thumb, whereas at longer distances, fingers farther from the thumb had higher usable ratios. Additionally, the usable ratio for fingers was higher in the four DoF cases (Case 2 and 4) than in the three DoF cases (Case 1 and 3). For the thumb, the usable ratio was higher for five DoF cases (Case 2 and 4) at short distances to other fingers, while at longer distances, the four DoF cases (Case 1 and 3) were more frequently used. This is because the thumb has a different DoF configuration compared to the other fingers and the usable range of A/A motions for five DoF fingers is limited.

Since the distance between contact points used for detection was normalized by the total hand length, this ratio is appropriate for identifying overall trends. However, to obtain an exact detection ratio, the stroke between the thumb and each finger that can occur in a given posture should be set to its maximum range, and the computation should be performed individually.

\begin{figure*}[!t]
    \centering
    \includegraphics[width=0.99\textwidth]{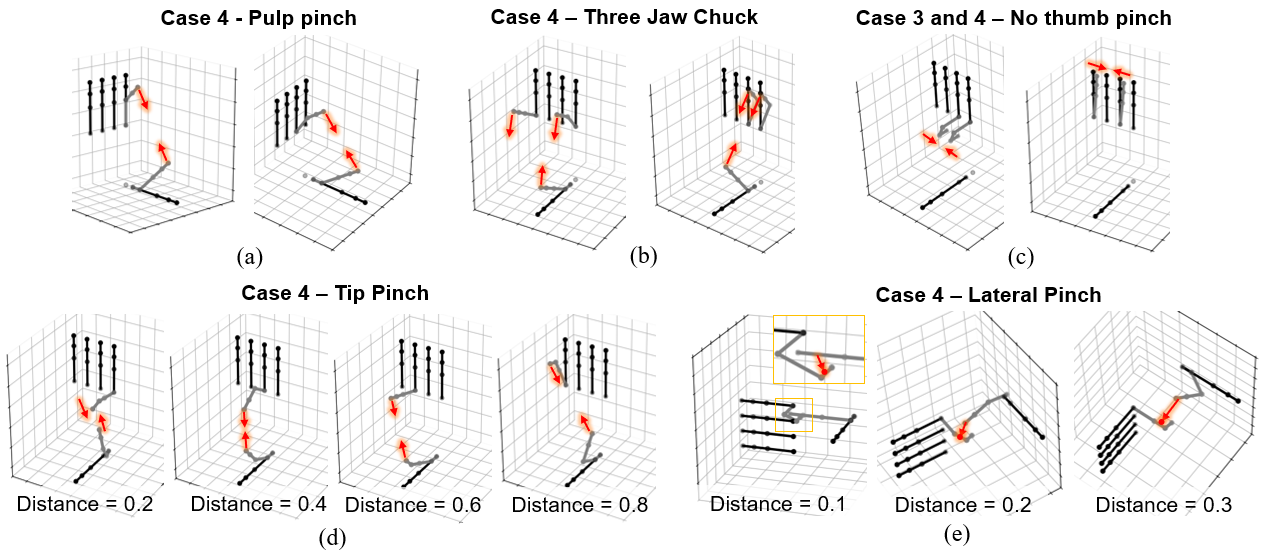}
    \caption{Example postures for different pinch types: (a) pulp pinch in Case 4, (b) three-jaw chuck pinch in Case 4, (c) no-thumb pinch in Cases 3 and 4, (d) tip pinch in Case 4, and (e) lateral pinch in Case 4}
    \label{fig:14}
\end{figure*}

When the pinch configurations evaluated using the three assessment methods are categorized into pulp pinch, three-jaw chuck pinch, no-thumb pinch, tip pinch, and lateral pinch, representative postures for each type are illustrated in Figure~\ref{fig:14}. The example postures are based on Case 4. Specifically, Figure~\ref{fig:14}(a) shows the pulp pinch, (b) the three-jaw chuck pinch, (c) the no-thumb pinch, (d) the tip pinch, and (e) the lateral pinch. For the pulp pinch cases shown in Figure~\ref{fig:14}(a), (b), and (c), no distance-based classification is applied. In contrast, for the tip pinch in Figure~\ref{fig:14}(d), the configurations are extracted by considering the distance between the thumb and each finger. The example distances are set to 0.2 for the index finger, 0.4 for the middle finger, 0.6 for the ring finger, and 0.8 for the little finger. For the lateral pinch shown in Figure~\ref{fig:14}(e), the phalangeal segments of the index finger corresponding to the thumb are indicated, and the representative distances are set to 0.1, 0.2, and 0.3 from left to right.

\subsection{Results}

The results are briefly summarized for each pinch evaluation method as follows. Tip pinch configurations are highly sensitive to the alignment between the thumb and the fingertip. Since the contact region is limited to the fingertip area, even small deviations in distal phalanx orientation can prevent stable pinch interaction. This indicates that robotic hands performing tip pinch require precise fingertip positioning and orientation control. In contrast, pulp pinch configurations allow a larger contact area between the thumb and finger pads. As a result, pulp pinch is less sensitive to small variations in distal phalanx alignment and enables stable pinch interactions under a wider range of finger configurations. Unlike tip and pulp pinch, lateral pinch involves contact between the thumb pad and the lateral surface of the finger. Therefore, the orientation of the finger side plays an important role in enabling stable pinch interactions.

Based on the kinematic structures determined by the DoF configurations of the hand, the pinch capabilities of the five-fingered hand were evaluated for four different cases. The evaluation results for the pinch types did not exhibit the same tendency across the four cases, and different results were obtained depending on the case. In detail, a configuration with a larger number of DoF can include the motions of a configuration with fewer DoF, and therefore it has the potential to produce a larger detected region. However, when the motion ranges differ, it cannot be guaranteed that a kinematic structure with more DoF will always have a larger detected region. Although the expansion of the workspace through an increase in DoF within the same specifications is an important factor, the region in which pinch operations can be performed may vary depending on the kinematic structure. Therefore, by using the evaluation method, the feasible regions for pinch operations of various kinematic structures can be examined in advance during the design stage.

The proposed evaluation framework enables systematic analysis of whether a given robotic hand kinematic structure can realize different pinch configurations. By identifying feasible and infeasible configurations, the evaluation results provide useful feedback during the design process of robotic hands. In particular, the framework allows designers to examine how variations in kinematic structure influence the feasibility of tip, pulp, and lateral pinch interactions. This capability makes it possible to analyze and compare different hand structures during the design stage before physical implementation.

%%%%%%%%%%%%%%%%%%%%%%%%%%%%%%%%%%%%%%%%%%%%%%%%%%%%%%%%%%%%%%%%%%%%%%
\section{Discussions}

In this study, a method was proposed to evaluate the feasibility of various pinch configurations according to the kinematic structure of a robotic hand. Previous studies have classified human pinch types from a functional perspective; however, systematic evaluation methods for determining whether such pinch configurations can be realized in robotic hands have not been sufficiently addressed. In this work, pinch configurations were evaluated based on three methods, and the results were analyzed according to the kinematic structure of the robotic hand.

The evaluation results show that tip pinch, pulp pinch, and lateral pinch require different geometric conditions. Furthermore, even in robotic hands with the same five-finger structure, the feasible regions for performing pinch configurations vary depending on differences in the kinematic structure arising from the DoF. In particular, it was observed that increasing the number of DoF does not necessarily result in a larger feasible region for pinch execution. This indicates that simply increasing the number of DoF does not guarantee diverse pinch interactions. These results suggest that, in robotic hand design, not only the number of DoF but also the arrangement of joints and the configuration of the kinematic structure significantly affect pinch configurations.

Therefore, the evaluation method can be used as a tool to examine, during the design process of a robotic hand, whether a specific kinematic structure is capable of performing various pinch interactions. It is useful for identifying which pinch configurations are structurally limited when conceptualizing and comparing different kinematic structures. Since the feasibility of pinch execution for various structures can be compared and analyzed prior to hardware implementation, the proposed approach is expected to serve as an evaluation framework that supports the decision-making process for determining the kinematic structure during the robotic hand design stage.

However, this study evaluated the feasibility of pinch configurations based on kinematic analysis, particularly focusing on a method intended for review during the design stage without considering specific objects or tasks. Therefore, factors that arise during real object manipulation—such as friction, compliance, and object geometry—should be appropriately considered in practical manipulation environments. In addition, since the evaluation was conducted based on the kinematic structure, design factors such as finger thickness and width were not reflected, which constitutes a limitation of this study. In future work, unlike the current approach that focuses on the design stage, we plan to consider evaluation methods that incorporate these design factors after the design has been completed. Furthermore, by reflecting the design of an actually implemented robotic hand, we aim to construct a set of candidate pinch tasks and derive feasible motion candidates that are reachable from the current state.

%%%%%%%%%%%%%%%%%%%%%%%%%%%%%%%%%%%%%%%%%%%%%%%%%%%%%%%%%%%%%%%%%%%%%%

\section{Conclusions}

An evaluation method was proposed to systematically analyze pinch configurations according to the kinematic structure of robotic hands. Pinch configurations were defined and evaluated based on the forms of tip pinch, pulp pinch, and lateral pinch observed in the human hand. Using the proposed method, pinch configurations were analyzed for five-finger robotic hands with different DoF configurations. The results show that the geometric conditions required for each pinch type differ, and that the range of feasible pinch configurations may vary even within the same hand structure. The analysis further indicates that not only the number of DoF but also the configuration of the kinematic structure plays a significant role in determining the feasibility of pinch configurations. This finding suggests that, in robotic hand design, simply increasing the number of DoF is not sufficient; the way in which the kinematic structure is configured also critically influences the ability to realize diverse pinch interactions. Such an approach enables the comparison and examination of candidate kinematic structures prior to actual hardware implementation and is expected to be applicable to a wide range of future robotic hand design studies.

\bibliographystyle{IEEEtran}
\bibliography{bibtex}

\end{document}